\documentclass[final]{cvpr}

\usepackage[utf8]{inputenc} % allow utf-8 input
\usepackage[T1]{fontenc}    % use 8-bit T1 fonts
\usepackage{url}            % simple URL typesetting
\usepackage{amsfonts}       % blackboard math symbols
\usepackage{nicefrac}       % compact symbols for 1/2, etc.
\usepackage{microtype}      % microtypography
\usepackage[english]{babel}
\usepackage{times}
\usepackage{epsfig}
\usepackage{graphicx}
\usepackage{tabularx}
\usepackage{booktabs}       % professional-quality tables
\usepackage{multirow}
\usepackage{multicol}
\usepackage{array}
\usepackage{dirtytalk}
\usepackage{adjustbox}
\usepackage{siunitx}
\usepackage{amsmath}
\usepackage{amssymb}
\usepackage{bm}
\usepackage[page]{appendix}

\usepackage[table]{xcolor}

\usepackage[pagebackref=true,breaklinks=true,colorlinks,bookmarks=false]{hyperref}

\def\cvprPaperID{****}
\def\confYear{CVPR 2021}

\begin{document}

%%%%%%%%% TITLE
\title{Neural Cellular Automata Manifold}

\author{%
  Alejandro Hernandez Ruiz \\
  IRI UPC \\
  Barcelona \\
%   \texttt{email} \\
  \and
  Armand Vilalta \\
  BSC UPC \\
  Barcelona \\
%   \texttt{email} \\
  \and
  Francesc Moreno-Noguer \\
  IRI UPC \\
  Barcelona \\
%   \texttt{email} \\
}

\maketitle

%%%%%%%%% ABSTRACT
\begin{abstract}
Very recently, the Neural Cellular Automata (NCA) has been proposed to simulate the morphogenesis process with deep networks. NCA learns to grow an image starting from a fixed single pixel. 
In this work, we show that the neural network (NN) architecture of the NCA can be encapsulated in a larger NN. This allows us to propose a new model that encodes a manifold of NCA, each of them capable of generating a distinct image. Therefore, we are effectively learning a embedding space of CA, which shows generalization capabilities. We accomplish this by introducing dynamic convolutions inside an Auto-Encoder architecture, for the first time used to join two different sources of information, the encoding and cell's environment information.
In biological terms, our approach would play the role of the transcription factors, modulating the mapping of genes into specific proteins that drive cellular differentiation, which occurs right before the morphogenesis.
We thoroughly evaluate our approach in a dataset of synthetic emojis and also in real images of CIFAR-10.
Our model introduces a general-purpose network, which can be used in a broad range of problems beyond image generation. 
\end{abstract}

\section{Introduction}

% GROWTH PROCESS
\begin{figure}[ht!]
\begin{centering}
    \adjincludegraphics[width=0.40\textwidth, trim={270 120 270 120}, clip]{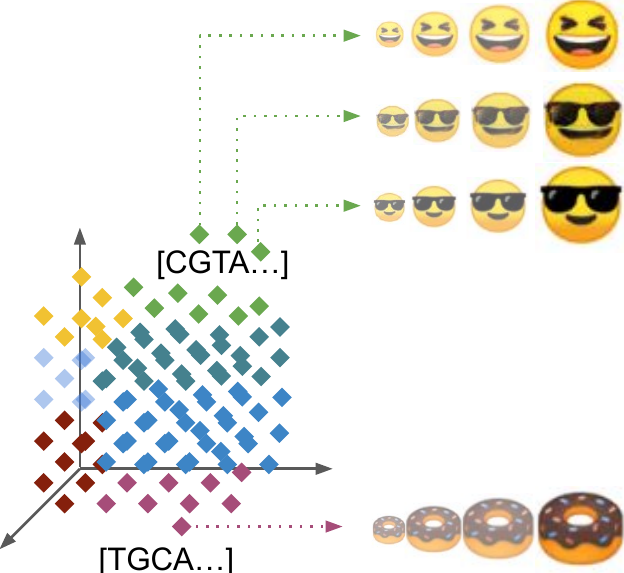}
    \caption{{\bf A manifold of Neural Cellular Automata.} Our model encodes a manifold of programs in a DNA-like encoding. The encodings are transformed into NCAs parameters, whose behavior is able to reconstruct (or ``grow'') a desired target image from a pixel seed.}
    \end{centering}
    \label{fig:teaser}
\end{figure}

Reproduction of multi-cellular organisms entails generating entire bodies from a single cell. Complex organisms also require to create different types of somatic cells and spatially arrange them to form the different tissues while ensuring temporal stability. These three aspects, cellular differentiation, morphogenesis and cell-growth control are the pillars of developmental biology. Computational methods are a key ingredient of the developmental biology study, up to the point that the term ``morphogene'' itself was coined by Turing~\cite{turing1952chemical} decades before its empirical demonstration. 

% In this context, there exist many different simulation techniques, including systems of partial derivative equation (PDEs), particle systems, and various types of Cellular Automata (CA)~\cite{zhang2015selforganizology, meinhardt2009algorithmic, peak2004evidence,jiao2011emergent,gonzalez2013dynamics}.

\begin{figure*}[ht!]
\begin{centering}
    \begin{tabular}{@{}c@{}}
    \includegraphics[width=\textwidth]{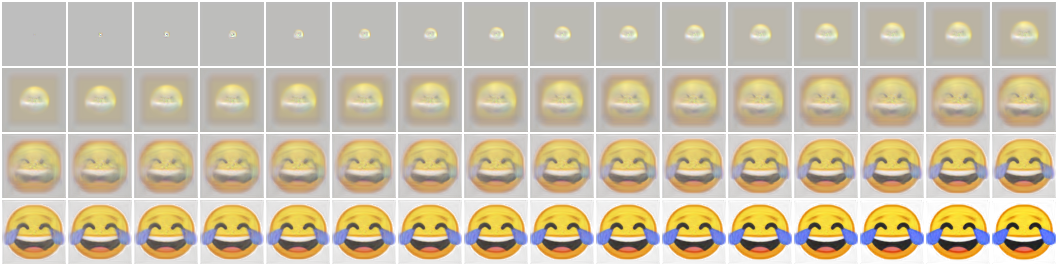}\\
    \includegraphics[width=\textwidth]{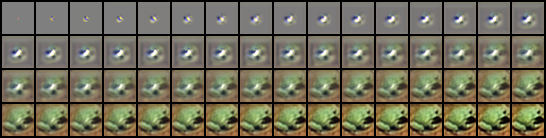}\\
    \end{tabular}
    \caption{{\bf Growth process} step by step from pixel seed image. Top, sample from  the full \texttt{NotoColorEmoji} dataset. Bottom, sample from the full \texttt{CIFAR-10} dataset.} 
    \label{fig:growth_process}
    \end{centering}
    \vspace{-2mm}
\end{figure*}

In this context, we model these fundamental processes using novel dynamic neural network architectures in combination with neural cellular automata (NCA)~\cite{mordvintsev2020growing}. We evaluate the proposed approach on the synthetic \texttt{NotoColorEmoji}~\cite{notoemoji} and real \texttt{CIFAR-10}~\cite{krizhevsky2009learning} image datasets. In both cases our model is able to ``grow'' the images with a low error (see Fig. \ref{fig:growth_process}).

% GROWTH PROCESS

% Although there is evidence of multi-cellular prokaryotes presenting differentiated cells~\cite{flores2010compartmentalized}, most complex organisms are eukaryotes (DNA is in a nucleus). Therefore, it seems that the existence of a protected genetic encoding is almost a requirement for the creation of complex organisms. This protection entails the use of a sophisticated transmission mechanism between the DNA and the cellular machinery, which proved to be related with the complexity of the organism~\cite{de2013transcription}. Transcription Factors (TF) are the main players in transcription regulation; coded by master regulatory genes, these factors activate or deactivate the expression of other genes in a regulatory cascade. 

\begin{figure*}[t!]
\begin{centering}
    \adjincludegraphics[width=\textwidth, trim={0 2.9cm 0 3.38cm}, clip]{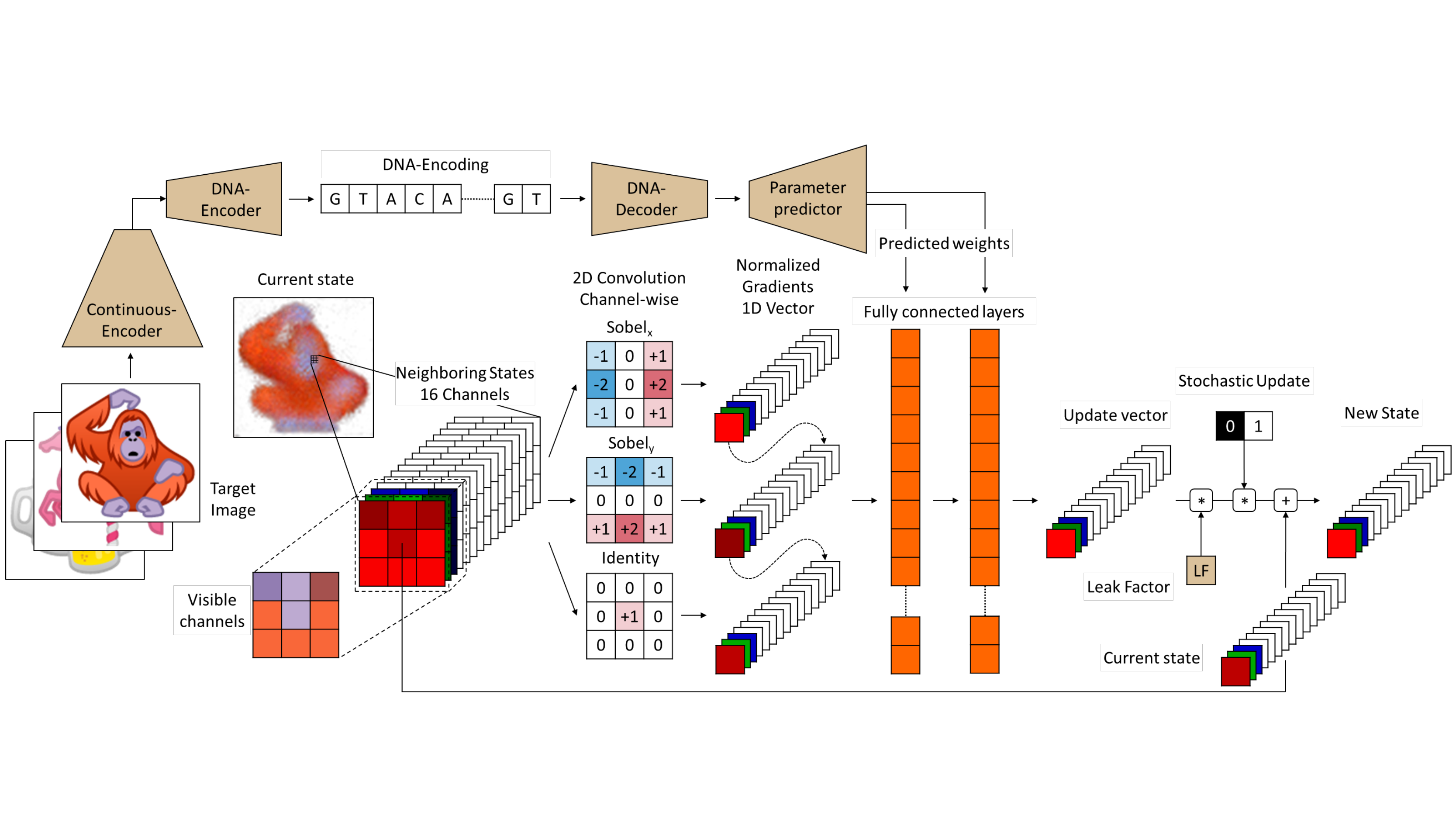}
    \caption{{\bf Model overview.} Beige elements contain trainable parameters while orange layers use only predicted parameters. See Fig.\ref{fig:ae_arch} for details of the architecture.}
    \label{fig:model}
    %\vspace{-4mm}
\end{centering}
\end{figure*}

Most complex organisms have their DNA protected inside a nucleus. DNA expression outside the nucleus is carried out by the cellular machinery and regulated via Transcription Factors (TF). Many TF are involved in defining an organism's spatial arrangement. Most of them are Morphogenes, soluble molecules that can diffuse and carry signals via concentration gradients. This biological model inspired our network architecture at the macro-scale (see Fig.~\ref{fig:model}). In our network, we create a vector encoding where common information for all cells is stored, as in DNA. The expression of such encoding by the ``cellular machinery'' is modulated by a Parameter Predictor (PP), with a similar role to that of TF. The ``cellular machinery'' (NCAM's Dynamic Convolutions) receives two sources of information: one from its DNA-encoding through the Parameter Predictor and another from the gradients of Morphogenes in its surrounding environment. Both are combined to model cell differentiation, producing the ``phenotype'' (color) and other Morphogenes (invisible channels) that drive the expression of neighboring cells.

In this work, we aim to learn not only a single model suitable for one specific  phenotype, but a single model suitable for different species with thousands of phenotypes.  For this purpose, we introduce a novel Auto-Encoder architecture with dynamic convolutions, which is  able to learn a embedding space of NCAs  suitable for thousands of images. The proposed model is trainable in an end-to-end manner,   and exhibits consistent generalization capabilities, in relation to the produced images.

%contrasts to the  original NCAs,  only suitable to generate a single image. 
%Our model builds upon two main ingredients, an Auto-Encoder arrchitettures trained end to end by back-propagation, and 
%That is, while a single NCA can generate a single image, in this paper we introduce a novel deep model able to learn  a representation space of NCAs, and which is suitable for thousands of images. To do so, we devise a 
%For this purpose, we use  an Auto-Encoder architecture trained end to end by back-propagation,  which learns vector representations of the NCA.% and exhibits generalization capabilities, in relation to the produced images.
%A single NCA can generate a single image, but our model learns a representation space of NCA, suitable for thousands of images. 
%Using an Auto-Encoder architecture trained end to end by back-propagation, our model learns vector representations of the NCA that exhibits generalization capabilities, in relation to the produced images. 

In our ``genetic engineering'' experiments (Sec.\ref{sec:experiments}) we show that it is possible to produce sensible new images from CA defined by new codes based on existing samples in the dataset. For the sake of similarity with the biological model, we also included an encoding conceptually similar to that of DNA, i.e. a vector encoding on a base of four possible categorical values similar to our DNA's cytosine, guanine, adenine, and thymine bases (the well-known ``CGAT'' sequences).

% To the best of our knowledge this is the first work to model a whole set of ``species'' with a neural network trainable end-to-end. Compared to the most similar approach \cite{mordvintsev2020growing}, which was able to learn to generate a single image, our single model is capable of reconstructing the whole 50.000 images of the \texttt{CIFAR-10} dataset. Moreover, previous restriction to work only on RGBA images has been removed boosting its applicability, first, to more abundant RGB images, but also to any kind of data in vectorial format.

In the biological domain, the model we propose could be useful to solve problems such as those faced by~\cite{peak2004evidence}, when it was not possible to completely model a plant's stomata complexity due to difficulties to obtain CA's function from noisy real data. Its applications can spread among many other biological domains where CA-modeling has been previously used, from tumor growth processes~\cite{jiao2011emergent} up to infection dynamics~\cite{gonzalez2013dynamics}.

Although biologically inspired, our model is capable of dealing with any vectorial data, making it useful in a broad range of problems. Having an embedding space of CA opens the possibility to create new CA that can generate unseen behaviors based on the recombination of features from learned CA samples.

The main contributions of this work are:
\begin{itemize}\setlength{\itemsep}{0pt}
    \item Introducing a new type of model that can learn a space of programs in the form of Cellular Automata, which are capable of producing desired target patterns.(see Sec.\ref{sec:nca_manifold})
    \item Showing that the learned program's space has generalization capabilities on the results the programs produce.(see genetic engineering experiments in Sec.\ref{sec:experiments})
    \item Showing that the embedding space is capable of learning and representing up to 50.000 different programs simultaneously with only 512 real-valued dimensions.(see CIFAR experiments in Sec.\ref{sec:experiments})
    \item Introducing a new fully dynamic network architecture, which generates CA's parameters from NN output.(see Sec. \ref{sec:dynamic_convs})
    \item The architecture proposed is trainable end-to-end, without need of pre-training any part, for datasets of different sizes and characteristics.  
    \item Extending the capabilities of the original NCA \cite{mordvintsev2020growing} from RGBA to RGB images, and potentially to any type of vectorial data.
    \item Demonstrate how to build a categorical embedding space similar to DNA, capable of encoding the same information that a continuous embedding space while achieving high robustness to random mutations.

\end{itemize}

%The main components of our the proposed architecture are the NCA, the dynamic convolutions and the Auto-Encoder. This combination allows us to create a manifold of CA, trainable end-to-end.

\section{Related}
\label{sec:related}

\vspace{1mm}
\noindent{\bf Cellular Automata} 
(CA) is a model based on a grid representation of the world. Each cell of the grid is an automaton or program which perceives the environment (i.e. its state and the state of the neighboring cells) and updates its state according to a fixed rule, generally a mathematical function. The rule is the same for every cell and does not change over time. CA models can be universal~\cite{cook2004universality}, which means that they can be programmed to emulate any other system without changing its underlying construction. Despite its expressive power, the main shortcoming of CAs is that given any specific rule, it is impossible to predict its behavior. This means that to understand its behavior, each rule has to be tried with multiple starting states, and its performance needs to be observed for many iterations. 
The Neural Cellular Automata (NCA), recently proposed by~\cite{mordvintsev2020growing}, is a class of Cellular Automata which uses an artificial neural network as update function, so the the NN parameters can be learned to obtain a desired behavior. Even more, the specific behavior is learned without prescribing the intermediate states, but instead by minimizing a loss towards a goal pattern, i.e. an image. 
NCA approach is conceptually different from other image generation techniques since it targets to create a complex program, which in turn is capable of generating the final image. Its characteristics are accordingly different, being, for instance, able to recover corrupted patterns. 
In this work we also address the requirement of a visibility channel of the original NCA \cite{mordvintsev2020growing}, allowing it a more general usability on RGB images, or any kind of data in vectorial form.
%Although it may superficially be similar to image generation techniques, it also has more complex characteristics, such as the ability to stay stable and regenerate the pattern if corrupted. This approach however has limited applications as it is, the main reason is because the NCA only learns a single pattern at a time, the second is because it is designed to work only with RGBA data, where the alpha channel has a special meaning. In our work, we address these issues and include an improved and more general NCA architecture, encapsulated inside our decoder network.

\vspace{1mm}
\noindent{\bf Conditioning Models.} 
%Several architectures have been proposed to augment CNNs capabilities, all performing some kind of scaling over intermediate representations between convolutional layers. 
%Attention models~\cite{xu2015show, wang2018non} scale the input according to a spatial probability distribution, being essentially a mask that tells the CNN which parts of the input are the most relevant.
Conditioning models use information either from the same network (e.g. squeeze and excitation block~\cite{hu2018squeeze}), from a joint network (e.g. style network~\cite{karras2019style, karras2020analyzing}) or from a completely independent source (e.g. language embedding~\cite{dumoulin2018feature-wise}), to scale the internal representation channel-wise. The key part is to inject into the network the all the information it needs to perform its task. The conditioning is not always considered an input in the standard manner, but rather transformation on it.
We could have used the conditioning approach to embed the NCA into our model, but this would have changed the original architecture by adding the conditioning layer to the CA. Such layer would also add an external source of information at each step, which would change the definition of a CA. Because our goal is to preserve as much as possible the architecture of a CA model, this approach was not ideal to us.

\vspace{1mm}
\noindent{\bf Dynamic Convolutions.} 
Instead of using conditioning to inject the information, we opted for the dynamic convolutions, in which the weights of the convolutional kernel are specifically computed for each sample. In previous works that use dynamic convolutions~\cite{klein2015dynamic, jia2016dynamic}, the architecture forks in two paths from the input image: one path computes the kernels while the other process the input through some fixed convolutions before feeding applying the previously computed dynamic kernels. In contrast, in our approach, the kernel weights are generated from an encoding which is completely independent (and different) from the image to be processed by them. Indeed, the image to be processed by our dynamic convolutions is the pixel seed, which is the same for all targets.

\vspace{1mm}
\noindent{\bf Dynamic Networks.} 
The formulation of our architecture makes the decoder a fully dynamic network, in which the weights are computed on the fly for each sample. In this perspective, it is related to the deep fried transform\cite{yang2015deep} one shot learners\cite{bertinetto2016learning} and the HyperLSTM\cite{ha2016hypernetworks}. The goal in these works was to classify images or text, which is a very different task from ours. 
To the best of our knowledge, our model is the only using a fully convolutional and fully dynamic network architecture in a generative setting. 
%To our knowledge, our model is unique in that it is a fully convolutional, fully dynamic network architecture for image generation. 
% More over, the proposed architecture makes no assumptions on the type of the data that is being generated, which implies that this model could be easily extended to handle any kind of data.
% Said that before.

\section{Neural Cellular Automata Manifold}
\label{sec:nca_manifold}

We already described in the introduction the biological inspiration of our model. Next, we move forward to its detailed formulation.
The Neural Cellular Automata Manifold (NCAM) uses dynamic convolutions to model a space of NCAs. This space is learned through an Auto-Encoder framework. Formally, our model is defined by the following equations:
\begin{equation} \label{eq:model}
\begin{split}
    \bm{I}^t & = \{\bm{C}_{ij}^{t}\} \quad \forall i,j \in \bm{I}\\
    \bm{C}_{ij}^{t} & = f(\bm{C}_{ij}^{t-1}, \bm{M}_{kl}^{t-1},\kappa(\bm{e}^I,\bm{\theta}), \theta_{LF}) \quad \forall (k,l) \in \bm{\epsilon}_{ij}\\
    \bm{M}_{ij}^{t} & = g(\bm{C}_{ij}^{t-1}, \bm{M}_{kl}^{t-1},\kappa(\bm{e}^I,\bm{\theta}) , \theta_{LF}) \quad \forall (k,l) \in \bm{\epsilon}_{ij} \\
    \kappa(\bm{e}^I, \bm{\theta}) & = \mathcal{P}(\mathcal{D}(\bm{e}^I,\bm{\theta}_{\mathcal{D}}),\bm{\theta}_{\mathcal{P}}) \\
    \bm{\theta} & = \{\bm{\theta}_{\mathcal{P}}, \bm{\theta}_{\mathcal{D}}\} \\
    \bm{\epsilon}_{ij} & = (\{i-n_x,\dotsc,i+n_x\}, \{j-n_y,\dotsc,j+n_y\}) \\
\end{split}
\end{equation}
where $\bm{I}^t$ is the image generated at step $t$, $\bm{C}_{ij}^{t}$ is the color vector (RGB or RGB-Alpha) of the pixel in position $(i,j)$, $\bm{M}_{ij}^{t}$ is the corresponding vector of ``Morphogenes'' (i.e. invisible channels in the grid), $\bm{\epsilon}_{ij}$ are indices of the neighborhood of the cell $(i,j)$ which extend $n_x, n_y$ positions to each side in $x$ and $y$ axis, and $\bm{e}^I$ is the vector encoding the image. $f(\cdot)$ and $g(\cdot)$ are the functions implemented as an NCA to predict the colors and ``Morphogenes'', respectively. $\kappa(\bm{e}^I, \bm{\theta})$ is the function that predicts the weights of the NCA from the encoding $\bm{e}^I$ and its learned parameters $\bm{\theta}$, which is the composition of the functions learned by the DNA-decoder $\mathcal{D}(\cdot)$ and the Parameter Predictor $\mathcal{P}(\cdot)$. The learned parameters are $\bm{\theta}_{\mathcal{P}}$, $\bm{\theta}_{\mathcal{D}}$ and $\theta_{LF}$, the Leak Factor (see Sec. \ref{sec:arch}).

In order to train this model, we could simply feed it with arbitrary codes, compute the reconstruction error for corresponding target images, and back-propagate the error to learn the parameters. However, we found it more sensible to learn embedding codes that can exploit similarities between images. We, therefore, decided to use an Auto-Encoder architecture~\cite{lecun1987modeles} at the macro level, which learns to map the set of inputs to a latent space and reconstructs them back. The Auto-Encoder consists of an Encoder and a Decoder (see Fig.~\ref{fig:ae_arch}). The Encoder is composed of two main components: a continuous encoder, that maps the input to a continuous variables encoding; and a DNA-encoder, that transforms this encoding to a categorical variables encoding. The Decoder's structure is symmetrical, mapping first the DNA-encoding to a continuous encoding which, in turn, feeds the parameter predictor block. From an auto-encoder perspective, the NCA should be considered the third part of the decoder since it is the responsible for finally providing the reconstructed images. From a NN perspective, we can see the NCA as a stack of Residual CNN blocks sharing weights or a ``Recurrent Residual CNN''.

\subsection{Architecture Details} 
\label{sec:arch}

\begin{figure*}[t!]
\begin{centering}
    \adjincludegraphics[width=\textwidth, trim={15 25 15 25}, clip]{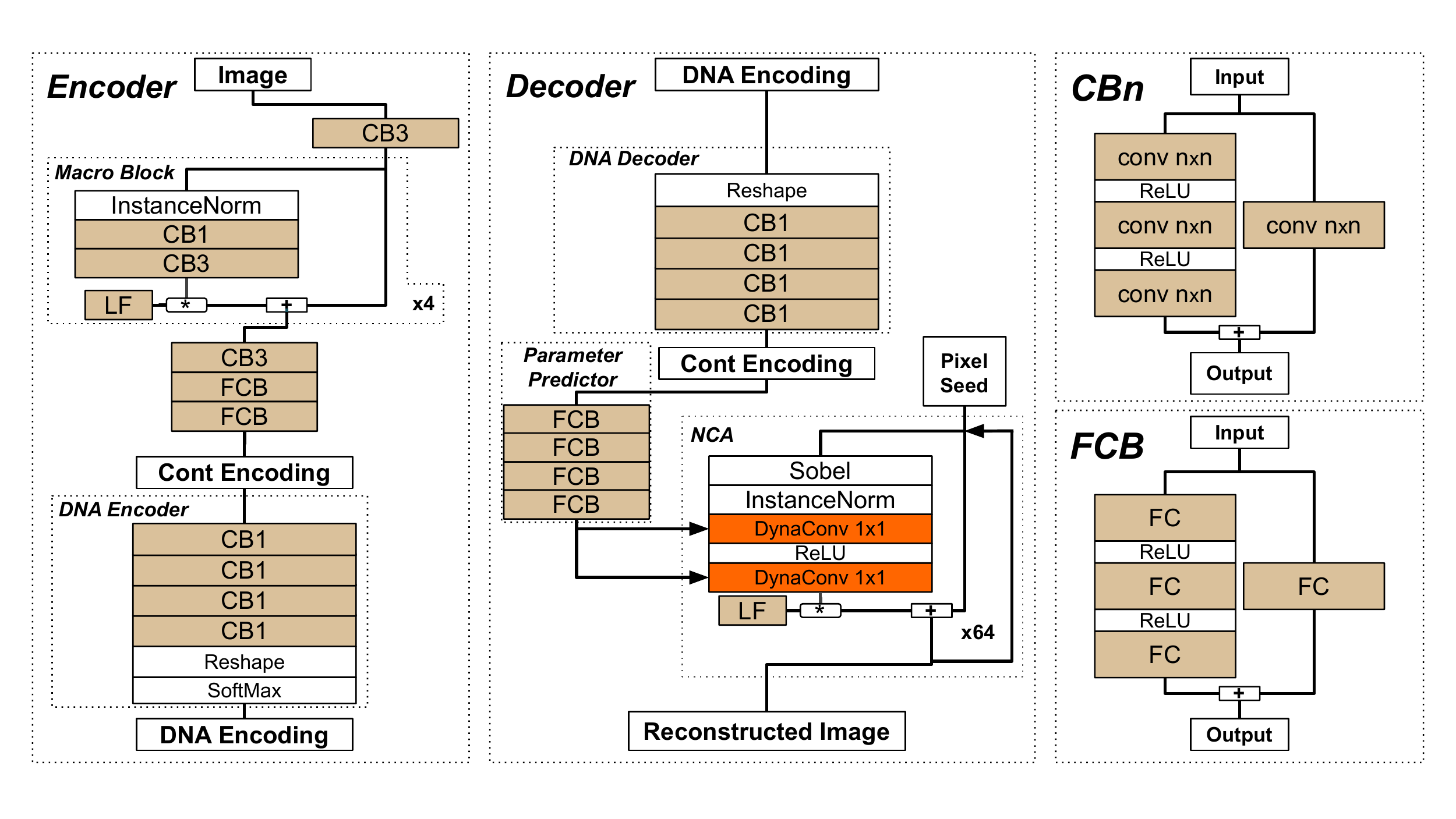}
    \caption{{\bf Architecture Details.} Beige elements contain trainable parameters while orange layers use only predicted parameters. CBn blocks can be CB1 or CB3. All blocks share the same number of filters except the last blocks whose output matches the embedding or parameters dimensionality.}
    \label{fig:ae_arch}
    % \vspace{-4mm}
\end{centering}
\end{figure*}

The architecture of the net is based on different types of residual blocks~\cite{he2016deep}. The smallest building blocks are of 3 types: Convolutional Block 3x3 (CB3), Convolutional Block 1x1 (CB1) and Fully Connected Block (FCB) (see details in Fig.~\ref{fig:ae_arch}). Unlike most of previous works~\cite{he2016deep, xie2017aggregated, szegedy2017inception}, the blocks   do not modify input/output representation dimensionality, neither in space resolution or number of filters. While these characteristics can be useful for classification problems, where information needs to be condensed through processing steps, our intuition is that this is not desirable in a reconstruction problem since detail information is lost.

A significant peculiarity of CB1 and CB3 is the expansion of the number features in the internal layer by a factor of 4 and 2, respectively. 
%This block can be seen as a per pixel fully connected architecture, with a 4x expansion in the hidden layer. 
We consider that this increase of dimensionality of the embedding space allows for a better disentanglement of the data manifold. Notice that CB1 needs to be used in combination with CB3 to introduce neighbourhood information. This detail is opposed to previous architectures~\cite{he2016deep, xie2017aggregated, szegedy2017inception} that reduce the number of filters in the inner layer of the block, while increasing dimensionality in the short-cut path.

A specific detail of the architecture is the use of a Leak Factor (LF) as an aid for training stability. LF is a single learnable parameter that regulates how much each successive step contributes to the output. A low value of LF encourages the network to retain most of the knowledge from previous steps, avoiding to distort the image too abruptly at any given step. After the network has learnt the basics of the problem this scenario is less likely, thus the network can learn to let more and more information to leak through each step. LF is constrained  between $10^{-3}$ and $10^{3}$, initialized at $10^{-1}$. %for a smooth starting.
 
Unlike many architectures that only take the spatial mean of the convolutional output tensor to feed FC layers~\cite{he2016deep, xie2017aggregated, szegedy2017inception}, we use 3 additional slices. Being $(c * h * w)$ the dimensions of the output tensor, a spatial mean over ${h, w}$ provides a $(c)$ dimensional encoding of the image. A mean over ${h}$ yields a $(c * w)$ dimensional vector, and doing similarly over the other 2 dimensions ($c$ and $w$), we obtain the input to our FCB of $dimension = c + (c * h) + (c * w) + (h * w)$.

\subsection{DNA-encoding}
The latent vector embedding can be interpreted as the source code that dictates the behavior on the cells of each of the NCAs. This definition inspired us to think of a possible alternative representation of such source code: each value in the latent vector can be encoded into a categorical base, making our encoding compatible to that of the DNA. Notice that a simple quantization on the continuous variable would provide an ordinal discrete variable instead of a DNA-like categorical one. This~encoding~is~dimensioned to handle the same 32 bits of information of the corresponding float variable, thus no additional bottleneck or expansion is introduced here.

As we can see in Fig.~\ref{fig:ae_arch}, the DNA-encoder is composed of 4 successive CB1 feeding a softmax layer to obtain the 4-categories DNA-embedding. The DNA-decoder follows a symmetrical structure, mapping back each \say{gene} to a continuous feature. These CB1 are all 1D convolutions, independently expanding each feature of the continuous encoding to a 16-features \say{gene}. This independent processing of different variables makes no assumption about the \say{meaning} of each category for one variable in relation to its \say{meaning} in other variables, so the actual \say{meaning} depends only on latter interaction between corresponding continuous variables.

In order to train our encoding, we add a biologically plausible noise replacing half of the letters by randomly drawn letters. Our intent is not to precisely mimic the mutation rate of DNA ($2.5 \times 10^{-8}/generation$~\cite{nachman2000estimate}) but to enforce a high redundancy in the encoding.
% more similar to the required to reduce transcription errors that can happen during cells lives.

% Recent work by \cite{eraslan2019single} shows how they use a Auto-Encoder to denoise real RNA data. They use the Auto-Encoder to fill in gaps in the data that lead to incorrect interpretations in gene expression. This tells us that our conceptual model is not too far away from actual biology.
% Link: https://www.nature.com/articles/s41467-018-07931-2.pdf?origin=ppub

%GROWTH
\begin{figure*}[t]
\begin{centering}
    \begin{tabular}{@{}b{50pt}<{\centering}@{\hspace{5pt}}b{340pt}<{\centering}@{}}
        Target & \raisebox{-.4\height}{\adjincludegraphics[width=340pt, trim={198 0 0 0}, clip]{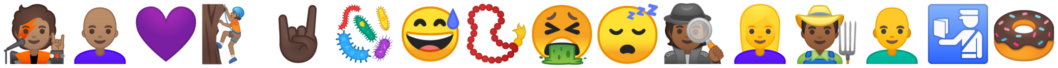}} \\
        \hline 
        CE-STO 16-512 & \raisebox{-.15\height}{\adjincludegraphics[width=340pt, trim={198 0 0 0}, clip]{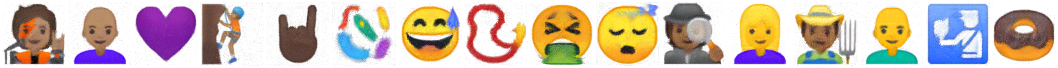}} \\
        DNA-STO 16-512 & \raisebox{-.15\height}{\adjincludegraphics[width=340pt, trim={198 0 0 0}, clip]{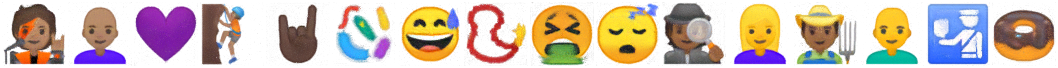}} \\
        
        CE-SYN 16-512 & \raisebox{-.15\height}{\adjincludegraphics[width=340pt, trim={198 0 0 0}, clip]{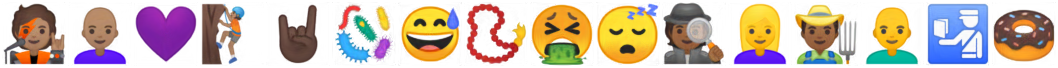}} \\
        DNA-SYN 16-512 & \raisebox{-.15\height}{\adjincludegraphics[width=340pt, trim={198 0 0 0}, clip]{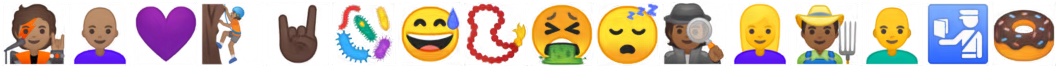}} \\
        
        DNA-STO 16-256 & \raisebox{-.15\height}{\adjincludegraphics[width=340pt, trim={198 0 0 0}, clip]{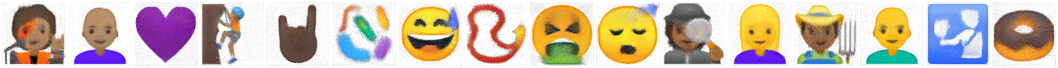}} \\
        DNA-STO 16-1024 & \raisebox{-.15\height}{\adjincludegraphics[width=340pt, trim={198 0 0 0}, clip]{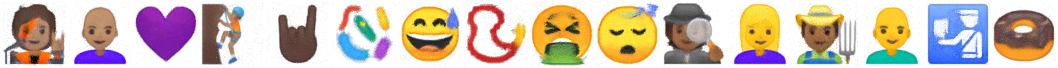}} \\
        
        %DNA-STO-NC32-ED512 & \adjincludegraphics[width=310pt, trim={198 0 0 0}, clip]{images/results/growth_DNA-STO-NC32} \\
        
    \end{tabular}
    \vspace{2mm}
    \caption{{\bf Growth results.} First row is a random set of images from the full \texttt{NotoColorEmoji} dataset. Following rows are generated by different variants of NCAM codified as:  CE: Continuous encoding, DNA: DNA-encoding; STO: Stochastic update, SYN: Synchronous update; 
    16-512: number of channels in NCA (16,32) -  Dimensionality of the continuous embedding (256, 512, 1024) (DNA dimensionality is 16 times that of continuous). }
    % All of them are normally trained except for DNA-STO-NC32-ED512 which has been trained with regeneration approach.} 
    \label{fig:growth_res}
    \vspace{-2mm}
\end{centering}
\end{figure*}

\subsection{Dynamic Convolutions} \label{sec:dynamic_convs}
Similarly to~\cite{klein2015dynamic}, for an image sample $I$, being $\bm{X}^I$  the input tensor of the convolution and $\bm{Y}^I$ the output tensor, we define the dynamic convolution as $\bm{Y}_{n}^I =\nolinebreak \sum_{m} \kappa(\bm{e}^I)_{mn} \star\nolinebreak \bm{X}_{m}^I$, where $\kappa(\bm{e}^I)_{mn}$ is the convolution kernel dynamically computed from the encoding for $I$, and $(m, n)$ are the input and output channels of the convolution. Although dynamic convolutions are quite a novel idea, in our case they are the simplest way to achieve our goal, keeping the original NCA architecture unchanged but encapsulated in a meta-learning system.

\subsection{Neural Cellular Automata Architecture}
\label{sec:nca}
The last component of the decoder is an NCA. To reconstruct an image, the NCA starts from a pixel seed image (typically a blank image with a different pixel). Then, each cell in the grid recurrently perceives the environment and computes an update. After a fixed number of steps, the output is evaluated and the reconstruction error back-propagated. In~\cite{mordvintsev2020growing}, NCA's kernel weights are directly updated, but in our case the error is back-propagated further, through the NCA and to the parameter predictor, updating its weights instead of the NCA's.
We can divide the NCA architecture in two main blocks, perception and update. 

The \textbf{Perception Layer} was designed to propagate the gradients across the grid in the 16 channels composing it (the first four corresponding to the visible RGBA). This is achieved with manually defined Sobel filters (see Fig.~\ref{fig:model}), similarly to~\cite{mordvintsev2020growing}. Since our network's architecture has only $ReLU$ activation functions, the activations have an unbounded positive value. Given the recurrent architecture of the NCA, during the training process, an exponential reinforcement can occur, leading activations and weights to $\infty$ and degenerate states. To avoid this problem, we apply instance normalization~\cite{ulyanov2016instance} on top of the Sobel filters.

To compute the \textbf{Update}, we use the NCA architecture, consisting on 2 pixel-wise dense layers (implemented as 1x1 convolutions). The update is modulated by the Leak Factor (the only trainable parameter in our NCA), analogous to the LF used in the Continuous Encoder. Finally, the update is stochastically applied on the current state with an independent probability of $p=0.5$ for each cell.

\section{Experiments}
\label{sec:experiments}

% We next present several experiments to highlight different aspects of our Neural Cellular Automata Manifold (NCAM). We also provide qualitative and quantitative evaluation in MSE.

\vspace{1mm}
\noindent{\bf Datasets.} \noindent\texttt{NotoColorEmoji} dataset~\cite{notoemoji}: 2924 images of synthetic emojis. This kind of images are very simple and with sharp edges, therefore it is relatively easy to visually assess the quality of the produced images. The original images are 128x128 pixels but in our experiments we downscaled them to 64x64 to reduce the computational requirements. \texttt{CIFAR-10} dataset~\cite{krizhevsky2009learning}: 50.000 real 32x32 images from 10 categories: plane, car, bird, cat, deer, dog, frog, horse, ship, truck. The number of images and its variability make it challenging.

\vspace{1mm}
\noindent{\bf Continuous vs. DNA encoding.} Generation results for these two sets of experiments are very similar, both visually (see Fig.~\ref{fig:growth_res}, rows 1-2) and numerically in MSE (CE: $0.01591$, DNA: $0.01633$). Results show no clear reduction or increase on performance with the extra processing involved in the DNA encoding and decoding process. We consider that the proposed methodology achieves the desired objective of obtaining a categorical encoding equivalent to the continuous encoding commonly used in DNNs. Moreover, we applied a 50\% chance of category error during the tests to showcase the remarkable robustness achieved by this encoding.

%GROWTH ON REDUCED DATASETS
\begin{figure*}[t!]
\begin{centering}
    \begin{tabular}{@{}b{40pt}@{\hspace{5pt}}b{350pt}<{\centering}@{}}
     
        \multirow{2}{*}{\texttt{Chars}} & \adjincludegraphics[width=350pt, trim={0 0 0 0}, clip]{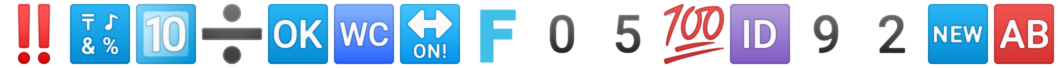} \\
        & \adjincludegraphics[width=350pt, trim={0 0 0 0}, clip]{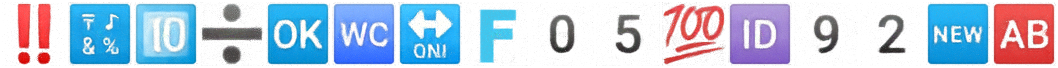} \\
     
        \multirow{2}{*}{\texttt{Emos}} & \adjincludegraphics[width=350pt, trim={0 0 0 0}, clip]{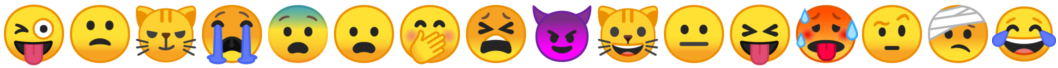} \\
        & \adjincludegraphics[width=350pt, trim={0 0 0 0}, clip]{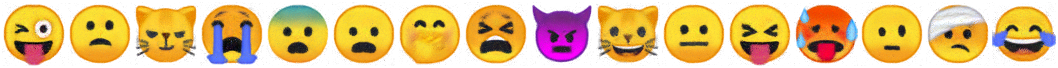} \\
     
        \multirow{2}{*}{\texttt{Heads}} & \adjincludegraphics[width=350pt, trim={0 0 0 0}, clip]{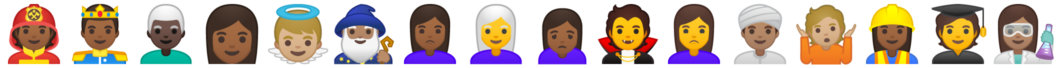} \\
        & \adjincludegraphics[width=350pt, trim={0 0 0 0}, clip]{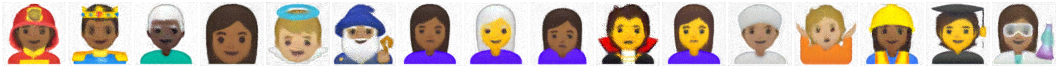} \\
     
        \multirow{2}{*}{\texttt{Variety}} & \adjincludegraphics[width=350pt, trim={0 0 0 0}, clip]{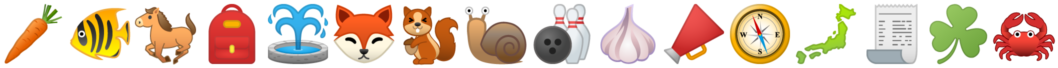} \\
        & \adjincludegraphics[width=350pt, trim={0 0 0 0}, clip]{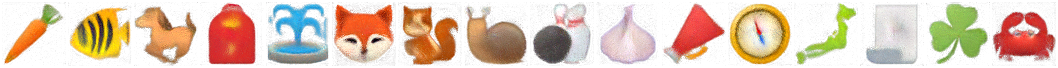} \\

    \end{tabular}
    \caption{{\bf Growth results on reduced datasets.} For each subset of \texttt{NotoColorEmoji}, first row are random images while the second are generated using DNA-encoding and stochastic update.} 
    \label{fig:data_res}
    \vspace{-2mm}
\end{centering}
\end{figure*}

\vspace{1mm}
\noindent{\bf Stochastic vs. synchronous update.} We consider that the stochastic update feature of the NCA, while being relevant in the biological analogy, may suppose an extra difficulty in the learning process. In these experiments we remove the stochastic update ($p=0.5$) making it synchronous at every time step ($p=1.0$). Notice that this change implies, on average, doubling the number of steps on the NCA processing. It also makes more reliable the expected state of neighboring cells. This modification reduced the error significantly in both scenarios, continuous and DNA encoding. Note that images generated with this approach succeed in reconstructing even the finest details (see Fig.\ref{fig:growth_res}, rows 3-4). MSE is one order of magnitude below stochastic approaches (MSE: CE: $0.00199$, DNA: $0.00262$). With either kind of update, it is remarkable the absence of artifacts in the background of the emojis, given that we removed the alive masking mechanism used in~\cite{mordvintsev2020growing}, proving that our design is able to learn by itself the limits of the figure.

\vspace{1mm}
\noindent{\bf Effect of encoding dimensionality.} We next evaluate our method under significant changes in the encoding dimensionality setting it to half (256) and double (1024) size. Notice that these dimensionalities refer to the continuous encoding, the actual DNA-encoding dimensions are 16 times larger. The experiments show (see Fig. \ref{fig:growth_res}, rows 5-6) that there is a significant quality degradation when the dimensionality is reduced to half while there is no appreciable improvement when it is doubled (MSE: 256: $0.02302$, 1024: $0.01584$). Therefore, we consider that the embedding size  is not a bottleneck for the problem at hand.

% We also performed experiment increasing the other main parameter, the number of channels in the image. This has a direct impact on the information available for the NCA. It should be easier to create complex patterns when there are more morphogenic gradients available. Results show a clear increase in the training speed and the quality of the generated images. This increase in dimensionality help the method to overcome the difficulties of the stochastic approach, achieving results similar to those of the synchronous update. (MSE: 32: $0.01...$)

\vspace{1mm}
\noindent{\bf Smaller datasets.} In order to assess the challenge that the size of the dataset and its visual complexity poses on the proposed method, we experiment on 4 different subsets of the \texttt{NotoColorEmoji} dataset, classified by the authors according to their visual appearance. \texttt{Chars} (96 images): emojis containing characters of the Latin set and a few symbols. They are simple shapes, usually with straight lines and few colors. \texttt{Emos} (103 images): round yellow faces showing emotions. 
%Images of people or visually very different emojis have been discarded. 
\texttt{Heads} (858 images): images of person heads, typically showing different professions. All theses images include the full set of versions according to skin tone. %Heads in different scales or strange skin tones have been discarded in sake of uniformity. 
\texttt{Variety} (684 images): images not present in other sets that are also visually different among them. It mainly includes animals, objects and food.
%Variations over the same base image (eg. clocks, moon phases) have been removed.

Results show that, as expected, problems have a growing level of difficulty in terms of MSE: \texttt{Chars}: $0.00976$ < \texttt{Emos}: $0.01799$ < \texttt{Heads}: $0.02439$ < \texttt{Variety}: $0.05035$ which can also be visually assessed on Fig.~\ref{fig:data_res}. The primary factor is the visual complexity or variety, not the size of the dataset. Linear simple shapes seem easier to generate than more rounded or intricate ones.

% GENETIC ENGINEERING
\begin{figure}[t]
\setlength{\tabcolsep}{3pt}
\begin{centering}
\begin{tabular}{cccccccc}
    & $mean$ &
    \multicolumn{2}{c}{\raisebox{-0.4\height}{\adjincludegraphics[height=22pt, trim={0 0 0 0}, clip]{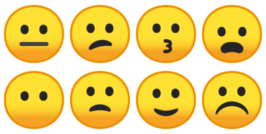}}} & &
    $mean$ &
    \multicolumn{2}{c}{\raisebox{-0.4\height}{\adjincludegraphics[height=22pt, trim={0 0 0 0}, clip]{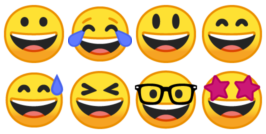}}}\\
    \cline{2-4} \cline{6-8}
    & $0.5$ & $0.7$ & $0.9$ & & $0.5$ & $0.7$ & $0.9$ \\
    \multicolumn{1}{c|}{} & \adjincludegraphics[width=22pt, trim={0 0 0 0}, clip]{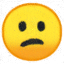} &
    \adjincludegraphics[width=22pt, trim={0 0 0 0}, clip]{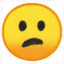} &
    \adjincludegraphics[width=22pt, trim={0 0 0 0}, clip]{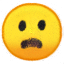} & &
    \adjincludegraphics[width=22pt, trim={0 0 0 0}, clip]{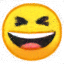} &
    \adjincludegraphics[width=22pt, trim={0 0 0 0}, clip]{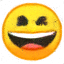} &
    \adjincludegraphics[width=22pt, trim={0 0 0 0}, clip]{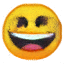} \\
    \hline
    \multicolumn{1}{c|}{\adjincludegraphics[width=22pt, trim={0 0 0 0}, clip]{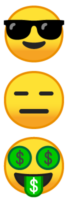}} &
    \adjincludegraphics[width=22pt, trim={0 0 0 0}, clip]{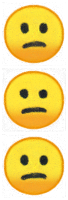} &
    \adjincludegraphics[width=22pt, trim={0 0 0 0}, clip]{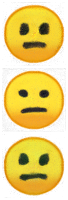} &
    \adjincludegraphics[width=22pt, trim={0 0 0 0}, clip]{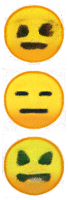} & &
    \adjincludegraphics[width=22pt, trim={0 0 0 0}, clip]{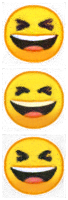} &
    \adjincludegraphics[width=22pt, trim={0 0 0 0}, clip]{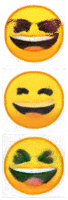} &
    \adjincludegraphics[width=22pt, trim={0 0 0 0}, clip]{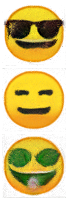} \\
\end{tabular}\\
\vspace{2mm}
\end{centering}
    \caption{{\bf ``Genetic engineering'' results.} On top: source images for each of two mean encodings. Just below: mean images generated by different thresholds (0.5, 0.7, 0.9): the higher, more common need to be the ``genes''. Three bottom rows: On the left: 3 original target images; On the right: different images generated when mean ``genes'' are injected to targets.}
    \label{fig:genetic_engineering}
    \vspace{-4mm}
\end{figure}

%CIFAR 10
\begin{figure*}[t!]
\begin{centering}
    \begin{tabular}{@{}b{50pt}<{\centering}@{\hspace{5pt}}b{340pt}<{\centering}@{}}
        Target & \raisebox{-.45\height}{\adjincludegraphics[width=340pt, trim={136 0 68 0}, clip]{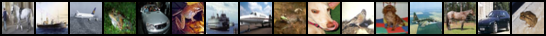}} \\
        \hline 
        
        CE-SYN 16-512 & \raisebox{-.2\height}{\adjincludegraphics[width=340pt, trim={136 0 68 0}, clip]{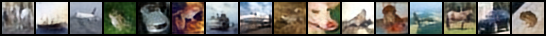}} \\
        DNA-SYN 16-512 & \raisebox{-.2\height}{\adjincludegraphics[width=340pt, trim={136 0 68 0}, clip]{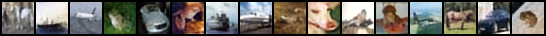}} \\
        
    \end{tabular}
    \vspace{2mm}
    \caption{{\bf CIFAR-10 results} using the best methods: Continuous and DNA Encoding with Syncronous update and baseline dimensionalities.}  
    \label{fig:cifar_res}
    \vspace{-2mm}
\end{centering}
\end{figure*}

\vspace{1mm}
\noindent{\bf Ablation Study.}
A small ablation study using the \texttt{variety} dataset gives us the following results: Our proposed architecture achieves a MSE of $0.0504$. w/o leak factor: $0.0597$. w/o normalization $0.0627$. w/o slices: $0.0670$. These numbers could be further improved by training longer and tuning the training parameters. In our preliminary experiments we observe that the tendency of the error does not change, thus the ordering of the different options for the model remains the same over time.

\vspace{1mm}
\noindent{\bf Comparison to NCA.}
The method defined in \cite{mordvintsev2020growing} is a qualitative reference but it can not be a baseline since the original NCA model was designed to tackle a different problem, and the learned CA can only produce a single image. Nevertheless, using the code\footnote{\url{https://colab.research.google.com/github/google-research/self-organising-systems/blob/master/notebooks/growing_ca.ipynb}} provided by the authors to train a $64\times64$px CA, we obtain a MSE in the order of $0.001$, comparable to our results with continuous encoding, but on a single image.

\vspace{1mm}
\noindent{\bf \say{Genetic engineering}.} %We perform some experiments on the DNA-encodings. 
We next play a little of \say{genetic engineering}, injecting part of the DNA-encoding from some images to others. To do so, we first generate a mean encoding of a group of source images that share a visual trait. We compute the mean of the 4 discrete features over the samples (i.e. DNA categories) and take the ones with values over a defined threshold. Notice that since they are produced by a softmax we can not have two values over 0.5. If none of the values is over the threshold we would have new category \say{none} which the DNA-decoder will ignore. With this encoding we generate the mean image. The parts of the mean encoding that are not \say{none} will substitute corresponding parts of target image encoding.

Results in Fig.~\ref{fig:genetic_engineering} show some interesting properties of the embedding. The lack of several features in the mean encoding causes no perturbation in the common traits and usually also provides a reasonable solution for the uncommon. The transfer process does not disrupt other common traits in target images. If the traits transferred collide with existing traits, they produce mixed shapes. We can observe that some traits that are not common in the source images are also transferred, suggesting some kind of trait dominance.

\vspace{1mm}
\noindent{\bf CIFAR-10 results.} Finally, we report results on \texttt{CIFAR-10} dataset~\cite{krizhevsky2009learning}. Note that in this case,  our NCAM model is capable to generate up to 50K different images. In Fig.~\ref{fig:growth_process}  we show an example of image generation for this case. 
%in Fig.~\ref{fig:growth_process} we can see an example of image generation for \texttt{CIFAR-10} dataset~\cite{krizhevsky2009learning}. 
On this dataset we only experimented with the synchronous update (best solution) since it is difficult to appreciate the level of detail required to visually evaluate the quality of the results (see Fig.~\ref{fig:cifar_res}). However, MSE values obtained are CE: $0.00717$ DNA: $0.00720$, perfectly comparable with those of \texttt{NotoColorEmoji}, which implies that the NCAM model also has enough capacity for this dataset.

%Stochastic update makes the model more robust to perturbations, gaining some regenerative properties. As the dropout in classical architectures this forces the model to behave like an ensemble model, in this case it would create an ensemble between the channels.

\section{Conclusions}

Machine learning techniques are already considered critical for the study of epigenetic processes, meddling between genetic information and their expression, which hold immense promise for medical applications~\cite{holder2017machine}. The model proposed here successfully simulates the main components of developmental biology, from DNA encoding to morphogenesis, at a high conceptual level. In our experiments, it is capable of reproducing almost 3.000 different emoji images, with a great level of detail, and up to 50.000 real images of the \texttt{CIFAR-10} dataset. Given its unique structure, capable of combining DNA-encoded and environment information, demonstrated scalability and robustness to noise, we consider it has potential in modeling genetic expression problems.

It is important to notice that the properties of the manifold learnt by the proposed Auto-Encoder will depend on the loss function used for training. In our work, we use MSE loss to learn to reproduce original images with high fidelity. If we were to use an adversarial loss~\cite{goodfellow2014generative}, we likely would obtain visually plausible images of the same class but not an exact replica. We consider that the application of a reinforcement learning loss~\cite{mnih2015human} could allow to produce a model driven by a fitness metric, such model would then be similar to genetic evolution. These simple adaptations open unfathomable use possibilities for the NCAM.

As stated in the introduction, CA models are widespread in biology and we consider that the generalization capabilities shown by the proposed method can be of interest in many different fields, specially where the model needs to be universal and able to fit noisy data. Cellular automata models have been proposed to obtain predictions on disease spreading~\cite{zhong2009simulation,gonzalez2013dynamics,slimi2006spreadable} and, more recently, on the COVID 19 pandemic evolution~\cite{pokkuluri2020novel,fang2020many}. We consider our model could also be of special interest for these tasks due to the capabilities shown and the ease to adapt it to different problems.

This work is a first step on creating embedding spaces to represent not only static information but behavioural patterns, effectively programs. Moreover, our experiments generate programs for multiple agents that, interacting through the environment, achieve the desired collective outcome. Given that our model is capable of dealing with any kind of vectorial data, we do not foresee any theoretical limitation to learn complex programs, to perform all sorts of different tasks, only from data and an approximate loss function. Finally, thanks to the NCAM embedding space, we can generate programs that produce new sensible novel behaviours unseen in the data.

{
    \small
    \bibliographystyle{ieee_fullname}
    \bibliography{references}
}

\end{document}